\documentclass{article}
\usepackage{ijcai22}

\usepackage{times}
\usepackage{soul}
\usepackage{url}
\usepackage[hidelinks]{hyperref}
\usepackage[utf8]{inputenc}
\usepackage[small]{caption}
\usepackage{graphicx}
\usepackage{mathtools}
\usepackage{amsmath}
\usepackage{amsthm}
\usepackage{amssymb}
\usepackage{amsfonts} 
\usepackage{booktabs}
\usepackage{algorithm}
\usepackage{algorithmic}
\usepackage{multirow}
\usepackage[dvipsnames, svgnames, x11names]{xcolor}
\usepackage{xspace}
\usepackage{pifont}
\usepackage{paralist}
\usepackage{enumitem}
\usepackage{color, colortbl}
\usepackage{subcaption}

\definecolor{LightCyan}{gray}{0.9}

\definecolor{dodgerblue}{rgb}{0.12,0.565,1}
\definecolor{antiquefuchsia}{rgb}{0.57, 0.36, 0.51}

\newtheorem{problem}{Problem}

\newcommand{\prob}{{\sc WSAD}\xspace}

\title{Weakly Supervised Anomaly Detection: A Survey}

\author{
Minqi Jiang$^{1,\ast}$\and
Chaochuan Hou$^{1,\ast}$\and
Ao Zheng$^{1,\ast}$\and
Xiyang Hu$^{4,}$\thanks{Equal contribution.} \and \\
Songqiao Han$^1$\and
Hailiang Huang$^1$\and 
Xiangnan He$^2$\and
Philip S. Yu$^3$\and 
Yue Zhao$^4$
\affiliations
$^1$Shanghai University of Finance and Economics
$^2$University of Science and Technology of China
$^3$University of Illinois Chicago
$^4$Carnegie Mellon University\\
{\emails
jiangmq95@163.com,
houchaochuan@foxmail.com,
zheng-ao@outlook.com
\{han.songqiao,hlhuang\}@shufe.edu.cn, 
xiangnanhe@gmail.com,
psyu@uic.edu,
\{xiyanghu,zhaoy\}@cmu.edu
}
}

\begin{document}

\maketitle

\begin{abstract}
Anomaly detection (AD) is a crucial task in machine learning with various applications, such as detecting emerging diseases, identifying financial frauds, and detecting fake news. However, obtaining complete, accurate, and precise labels for AD tasks can be expensive and challenging due to the cost and difficulties in data annotation. To address this issue, researchers have developed AD methods that can work with \textit{incomplete}, \textit{inexact}, and \textit{inaccurate} supervision, collectively summarized as \underline{w}eakly \underline{s}upervised \underline{a}nomaly \underline{d}etection (WSAD) methods. In this study, we present the \textit{first} comprehensive survey of WSAD methods by categorizing them into the above three weak supervision settings 
% (i.e., incomplete, inexact, 
% and inaccurate supervision) and 
across four data modalities (i.e., tabular, graph, time-series, and image/video data).
% including 42 algorithms in 3 weak supervision settings across 4 data modalities. 
For each setting, we provide formal definitions, key algorithms, and potential future directions. To support future research, we conduct experiments on a selected setting and release the source code, along with a collection of WSAD methods and data.
% Anomaly detection (AD) is an important machine learning task with numerous applications, including emerging disease detection, financial fraud detection, and fake news detection. 
% % \mq{remove the examples of applications?}.
% % For these real-world applications,
% However, it is often impractical to access complete, exact, or accurate labels for AD tasks due to the cost and challenges in data annotation.
% % \mq{what about inexact, i.e., coarse-grained labels?}, and thus 
% % Uniquely, AD applications often face label quality issues, including 
% % incomplete, inaccurate, and inexact ground truth labels. In the last decade, 
% Over the years, researchers have proposed AD methods that can work with \textit{incomplete}, \textit{inaccurate}, and \textit{inexact} supervision (i.e., ground truth labels), which are collectively known as \underline{w}eakly-\underline{s}upervised \underline{a}nomaly \underline{d}etection (\prob) methods. In this work, we provide the first systematic \prob survey by including \nalgorithms algorithms in the 3 weak supervision settings across \nmodality data modalities. For each setting, we provide formal definitions, describe key algorithms, and identify promising future directions.
% To foster future studies, we conduct experiments for a selected setting, and release the corresponding code along with collected \prob methods and data.
% % we summarize open problems and promising future directions for \prob. 
\end{abstract}

% Intro
\section{Introduction}
Anomaly detection (AD) aims to identify deviant samples from the general data distribution, which has many important applications in healthcare, finance, security, etc \cite{Aggarwal2013a,zhao2019pyod}. However, AD applications face data quality issues such as \textit{incomplete}, \textit{inexact}, and \textit{inaccurate} labels \cite{han2022adbench},
% \textcolor{red}{coarse-grained labels?}, 
which can be caused by the high cost and low accuracy of human annotation \cite{zhao2022admoe}. Concrete examples include inconclusive readings from inappropriate sensors, human mistakes due to repetitive work, and high labeling costs in financial fraud detection. These situations are collectively referred to as \underline{w}eakly \underline{s}upervised \underline{a}nomaly \underline{d}etection (\prob), in which complete, exact, and accurate labels are not available.

% Unlike the research setting where one may access complete and accurate labels, real-world applications often suffer from data quality issues \cite{han2022adbench}, e.g., noisy and incomplete labels. This is often caused by the inaccuracy and high cost of human annotation \cite{zhao2022admoe}.

In this survey, we adapt the categorization of weakly supervised learning (WSL) \cite{zhou2018brief} into the context of AD, and define the following three categories of \prob:

% Following the categories in weakly supervised machine learning (WSML) \cite{zhou2018brief}, we collectively consider the above label quality issues in AD as \underline{w}eakly \underline{s}upervised \underline{a}nomaly \underline{d}etection (\prob). \prob is therefore categorized into the following three buckets, as shown in Table~\ref{tab:category}:

\begin{itemize}[noitemsep,topsep=0pt,parsep=0pt,partopsep=0pt]
    \item \textbf{AD with incomplete supervision} concerns the situation where only part of the input data is labeled (\S \ref{sec.incomplete}).
    % where in addition to a large amount of unlabeled data, one could get access to a limited number of labeled samples, which is often insufficient to train a good learner.
    \item \textbf{AD with inexact supervision} happens when 
    % refers to the scenario where 
    the detected labels are not as exact as they need to be, e.g., only coarse-grained (instead of fine-grained) label information is available due to the labeling cost (\S \ref{sec.inexact}). 
    \item \textbf{AD with inaccurate supervision}
    % . Inaccurate supervision 
    refers to the problem where the labels are corrupted with noise, which may be caused by machine mistakes and annotation errors (\S \ref{sec.inaccurate}). 
    % information is not always ground-truth, i.e., some label noises like annotation errors may occur in the model training stage.
\end{itemize}

\noindent \textbf{Key Challenges}. 
% Beyond WSL, 
\prob faces unique challenges, such as extreme data imbalance and limited label availability---making existing WSL methods fail and thus yield limited performance for \prob \cite{zhao2022admoe}.
% we thus cannot directly leverage
% supervised 
% or WSL 
% \mq{Supervised?}
% Weakly Supervised Learning)
% existing WSL methods for \prob. 
% In fact, the literature has shown that WSL methods yield limited performance on \prob \cite{zhao2022admoe}.
In addition, AD has diverse applications across different data modalities, including tabular, time-series, image and video, and graph data,  requiring distinct and specialized designs.
% We try to cover as many data modalities as possible; see Table \ref{table:all_works} for details.

\noindent \textbf{Organization}. We compile this survey to summarize \prob methods under all three settings across various data modalities; Table \ref{table:all_works} may serve as a high-level summary. Namely, we provide details of AD with incomplete supervision, inexact supervision, and inaccurate supervision in \S \ref{sec.incomplete}, \ref{sec.inexact}, and \ref{sec.inaccurate}, respectively. Each setting is associated with a formal definition, key algorithms, and open problems and future directions.
% Table \ref{table:all_works} summarizes the covered algorithms by types of weak supervision as well as data modalities.
% With this in mind, we organize the paper as the pairs of \texttt{\{supervision type, data modality\}} (see Table \ref{tab:category} for pointers.)
% \input{tables/category.tex}
% NOT SURE, MAY BE A PROBLEM:
% \textcolor{orange}{However, since anomaly detection stems from an unsupervised learning paradigm (actually, data mining), it's properties particularly in different modalities make those problems unique and different from general machine learning. Specifically, those three major weakly-supervised settings are somewhat inter-changeable and the scenarios are restricted to the following.}
% Can we develop a real taxonomy for \prob?
% \mq{NLP modality? Voice? Speech?}
% 一个维度从数据特点,数据形态上面分(tabular..time-series..graph)
% 例如对于video而言,image提embedding,转化为类似于tabular data，结合时间维度,转化为类似于time-series data，如果考虑不同clip之间的联系（类似于考虑空间信息），类似于graph。所以NLP,voice,image属于不同的应用角度，tabular,time-series,graph可以看做不同的数据结构
% 第二个维度从应用角度(领域)总结, 例如NLP, voice, image

\noindent \textbf{Contributions}. We provide the first comprehensive survey on weakly supervised anomaly detection, summarizing existing methods into three weak supervision settings across four data modalities.
To foster future research, we release the corresponding
resources
% code and data 
at \url{https://github.com/yzhao062/wsad}.
% Our technical contributions are as follows:
% \begin{itemize}[noitemsep,topsep=0pt,parsep=0pt,partopsep=0pt]
%     \item \textbf{First systematic survey} on weakly supervised anomaly detection with detailed problem definitions.
%     \item \textbf{Comprehensive coverage} of \nalgorithms algorithms across \nmodality data modalities, along with selected empirical analysis.
%     \item \textbf{Open problems and future directions} for continued research on this important \prob problem.
% \end{itemize}

% \vspace{-0.05in}
\newcommand{\cmark}{\ding{51}}%
\newcommand{\xmark}{\ding{55}}%

\begin{table*}[ht]
\scriptsize
  \centering
\caption{Summary of \prob methods with categorizations by weak supervision type, backbone, and data modality. We define the abbreviation of the methods if there is no acronym. More details and the collected official code can be found at \url{https://github.com/yzhao062/wsad}. 
% \textcolor{red}{The order of the rows should match exactly their appearance in the paper. The key idea should match our subsection titles.}
% \mq{Should we point out the url of code and whether it is official? Such big table could refer to the ijcai survey paper "A Survey of Vision-Language Pre-Trained Models", "Few-Shot Learning on Graphs"}
}
\vspace{-0.1in}
\label{table:all_works}
\begin{tabular}{llccclc}
\toprule
% \textbf{Category}      & 
\textbf{Method} & \textbf{Reference} & \textbf{Venue} & 
\textbf{Backbone} & \textbf{Modalities} &       \textbf{Key Idea} & \textbf{Official Code}\\
% Key idea总结几个就可以，不要free text
\midrule
\rowcolor{LightCyan}
\multicolumn{7}{c}{\textcolor{black}{\textbf{\textit{Incomplete}} (\S \ref{sec.incomplete})}} \\
% \multirow{4}{*}{\textbf{\textit{Incomplete}} (\S \ref{sec.incomplete})}
% &   SNARE   &   \cite{mcglohon2009snare}  & Graph       &           & Belief propagation    \\
% &   AESOP   &   \cite{tamersoy2014guilt}  & Graph       &           & Belief propagation    \\
OE  &\cite{micenkova2014learning}       & KDD'14  &    -     & Tabular &  Anomaly feature representation learning & \xmark\\
XGBOD  &\cite{zhao2018xgbod}       & IJCNN'18  &   -      & Tabular &  Anomaly feature representation learning & \cmark\\
DeepSAD  &\cite{ruff2019deep}       & ICLR'20  & MLP        & Tabular &  Anomaly feature representation learning & \cmark\\
   ESAD      &\cite{huang2020esad}      &    Preprint      &      MLP      &   Tabular      &  Anomaly feature representation learning                  & \xmark\\
 DSSAD        &\cite{feng2021learning}   & ICASSP'21         &    CNN        & Image/Video        &          Anomaly feature representation learning          & \xmark\\
REPEN    & \cite{pang2018learning}  & KDD'18   & MLP        & Tabular &      Anomaly feature representation learning                   &  \xmark \\
AA-BiGAN & \cite{tian2022anomaly}   & IJCAI'22 & GAN        & Tabular &         Anomaly feature representation learning                & \cmark \\
Dual-MGAN      &\cite{li2022dual}         &      TKDD'22     &    GAN       &    Tabular     &     Anomaly feature representation learning                    & \cmark\\
 DevNet  & \cite{pang2019deepdevnet} & KDD'19   & MLP       & Tabular &   Anomaly score learning  &  \cmark\\
  PReNet & \cite{pang2019deepprenet} &  Preprint        & MLP       & Tabular &  Anomaly score learning                         &  \xmark   \\
  FEAWAD & \cite{zhou2021feature}   & TNNLS'21  & AE        & Tabular &     Anomaly score learning                       & \cmark    \\
SNARE    & \cite{mcglohon2009snare} & KDD'09    & -          & Graph   &  Graph learning and label propagation &   \xmark  \\
AESOP    & \cite{tamersoy2014guilt} & KDD'14    &  -         & Graph   &  Graph learning and label propagation  &   \xmark  \\
SemiGNN  & \cite{wang2019semi}      & ICDM'19   & MLP+Attention & Graph   &  Graph learning and label propagation & \xmark    \\
SemiGAD   & \cite{kumagai2021semi}   & IJCNN'21  & GNN       & Graph   &  Graph learning and label propagation  & \xmark \\
Meta-GDN & \cite{ding2021few}       & WWW'21    & GNN       & Graph   &   Graph learning and label propagation &  \cmark   \\
SemiADC  & \cite{meng2021semi}      & IS Journal'21 & GAN   & Graph & Graph learning and label propagation&  \xmark \\
SSAD     & \cite{gornitz2013toward} & JAIR'13   &  -         & Tabular &   Active learning                  &  \xmark   \\
AAD      & \cite{das2016incorporating} & ICDM'16 & -         & Tabular &  Active learning   &   \cmark  \\
SLA-VAE    & \cite{huang2022semi}      &   WWW'22        &    VAE      &     Time series    &          Active learning                    & \xmark \\
Meta-AAD & \cite{zha2020meta}       & ICDM'20   & MLP       & Tabular &  Reinforcement learning &   \cmark  \\
DPLAN    & \cite{pang2021toward}     & KDD'21   & MLP       & Tabular &  Reinforcement learning  &  \xmark   \\
GraphUCB & \cite{ding2019interactive} & WSDM'19 &   -        & Graph   &  Reinforcement learning  &  \cmark   \\

\midrule
\rowcolor{LightCyan}
\multicolumn{7}{c}{\textcolor{black}{\textbf{\textit{Inexact}} (\S \ref{sec.inexact})}} \\
% \multirow{4}{*}{\textbf{\textit{Inexact}} (\S \ref{sec.inexact})}
   % CMLF\mq{?}     &     \cite{hou2021stock} & CIKM'21 &  &     Time-series  &    Multi-granularity Data + Contrastive Learning        &      \\
%           &  \cite{georgescu2021anomaly}  &CVPR'21   &  CNN    &  Video  &   Multi-task Learning &   \cmark       \\

 MIL          &  \cite{sultani2018real}   & CVPR'18 & MLP          &  Video  &  Multiple Instance Learning  & \cmark        \\
TCN-IBL       &  \cite{zhang2019temporal} & ICIP'19  & CNN          & Video   &  Multiple Instance Learning  &  \xmark       \\
AR-Net        &  \cite{wan2020weakly}     &ICME'20   &  MLP         & Video   & Multiple Instance Learning   & \cmark        \\
 RTFM         &  \cite{tian2021weakly}    & ICCV'21  & CNN+Attention &  Video  &  Multiple Instance Learning  & \cmark            \\
Motion-Aware  &  \cite{zhu2019motion}     & BMVC'19  & AE+Attention& Video   &  Multiple Instance Learning  &   \xmark     \\
 CRF-Attention    &  \cite{purwanto2021dance} &ICCV'21   & TRN+Attention    &  Video  &  Multiple Instance Learning  &      \xmark   \\
MPRF         &  \cite{wu2021weakly}       &IJCAI'21  & MLP+Attention     &  Video  &  Multiple Instance Learning  &   \xmark      \\
 MCR         &  \cite{gong2022multi}      &ICME'22   &   MLP+Attention   & Video  &   Multiple Instance Learning &  \xmark        \\
  XEL          &   \cite{yu2021cross}      &  SPL'21  &  MLP          &  Video  &   Cross-epoch Learning       & \cmark         \\   
  MIST          &  \cite{feng2021mist}  &CVPR'21   & MLP+Attention     &  Video  &  Multiple Instance Learning  &    \cmark      \\
  MSLNet       &  \cite{li2022self}         & AAAI'22  & Transformer    &   Video &   Multiple Instance Learning &  \cmark        \\
 SRF             &  \cite{zaheer2020self}    & SPL'20   &  MLP          &  Video &  Self Reasoning &  \xmark            \\
  WETAS       &  \cite{lee2021weakly}     & ICCV'21  & MLP           &   Time-series/Video &   Dynamic Time Warping       &    \xmark     \\       
Inexact AUC   &  \cite{iwata2020anomaly}  & ML Journal'20  &  AE   &  Tabular  &  AUC maximization  &    \xmark      \\
  Isudra           &\cite{dahmen2021indirectly} &  TIST'21  &   -   &Time-series  &  Bayesian optimization  &   \cmark    \\

\midrule
\rowcolor{LightCyan}
\multicolumn{7}{c}{\textcolor{black}{\textbf{\textit{Inaccurate}} (\S \ref{sec.inaccurate})}} \\
% \multirow{2}{*}{\textbf{\textit{Inaccurate}} (\S \ref{sec.inaccurate})} 
 LAC &    \cite{zhang2021fraud}    & CIKM'21 &  MLP/GBDT  & Tabular         & Ensemble learning &    \xmark  \\
 ADMoE &    \cite{zhao2022admoe}    & AAAI'23 &  Agnostic  & Tabular         & Ensemble learning &    \cmark  \\
 BGPAD  &    \cite{dong2021isp}    & ICNP'21 & LSTM+Attention   &   Time series       & Denoising network &   \cmark   \\
 SemiADC  & \cite{meng2021semi}      & IS Journal'21 & GAN   & Graph & Denoising network&  \xmark \\
  TSN &    \cite{zhong2019graph}    & CVPR'19 & GCN   &   Video       &GCN  &   \cmark   \\
% LogLAB  &    \cite{wittkopp2021loglab}    & ICSOC'21 & Attention   &     Text     & PU Learning &   \cmark   \\
\bottomrule
\end{tabular}
\vspace{-0.1in}
\end{table*}%

% \comment{This survey paper has some discussion about deep weakly-supervised AD (see \cite{pang2021deep}, 8.2)} \\

\vspace{-0.1in}
\section{AD with Incomplete Supervision}
\label{sec.incomplete}
% \vspace{-0.05in}

% \subsection{Definitions and Real-world Applications}
\textbf{AD with incomplete supervision} refers to the situation where only part of the input data is labeled, where fully-supervised models do not apply. This setting is common in AD due to the high data labeling cost \cite{pang2021deep,han2022adbench}.
% due to the high cost of data annotations. 
For instance, accessing the complete label set of credit card fraud
% (where fraudulent activities are considered anomalies) 
requires annotating all the normal and abnormal transactions manually, which is often time-consuming and costly. Thus, most of the \prob literature focuses on the incomplete setting due to its universality.
% Notably, most of \prob works focus on this problem due to its universality.

\begin{problem}
[AD with incomplete supervision]
\textit{\em Given} an anomaly detection task with input data $\mathbf{X} \in \mathbb{R}^{n\times d}$
(e.g., $n$ samples and $d$ features) and partial labels $\mathbf{y}_{p} \in \mathbb{R}^{p}$ where $p < n$, 
train a detection model $M$ with $\{\mathbf{X}, \mathbf{y}_{p}\}$.
Output the anomaly scores on the test data $\mathbf{X}_\text{test} \in \mathbb{R}^{m\times d}$, i.e., $\mathbf{O}_\text{test} := M(\mathbf{X}_\text{test}) \in \mathbb{R}^{m\times 1}$.
% to achieve the best performance. 
% \mq{the definition of anomaly score like in ADBench?}
\label{problem:incomplete}
\end{problem}

\subsection{Key Algorithms}
\begin{figure}[!tp]
    \centering
    \includegraphics[width=8.6cm]{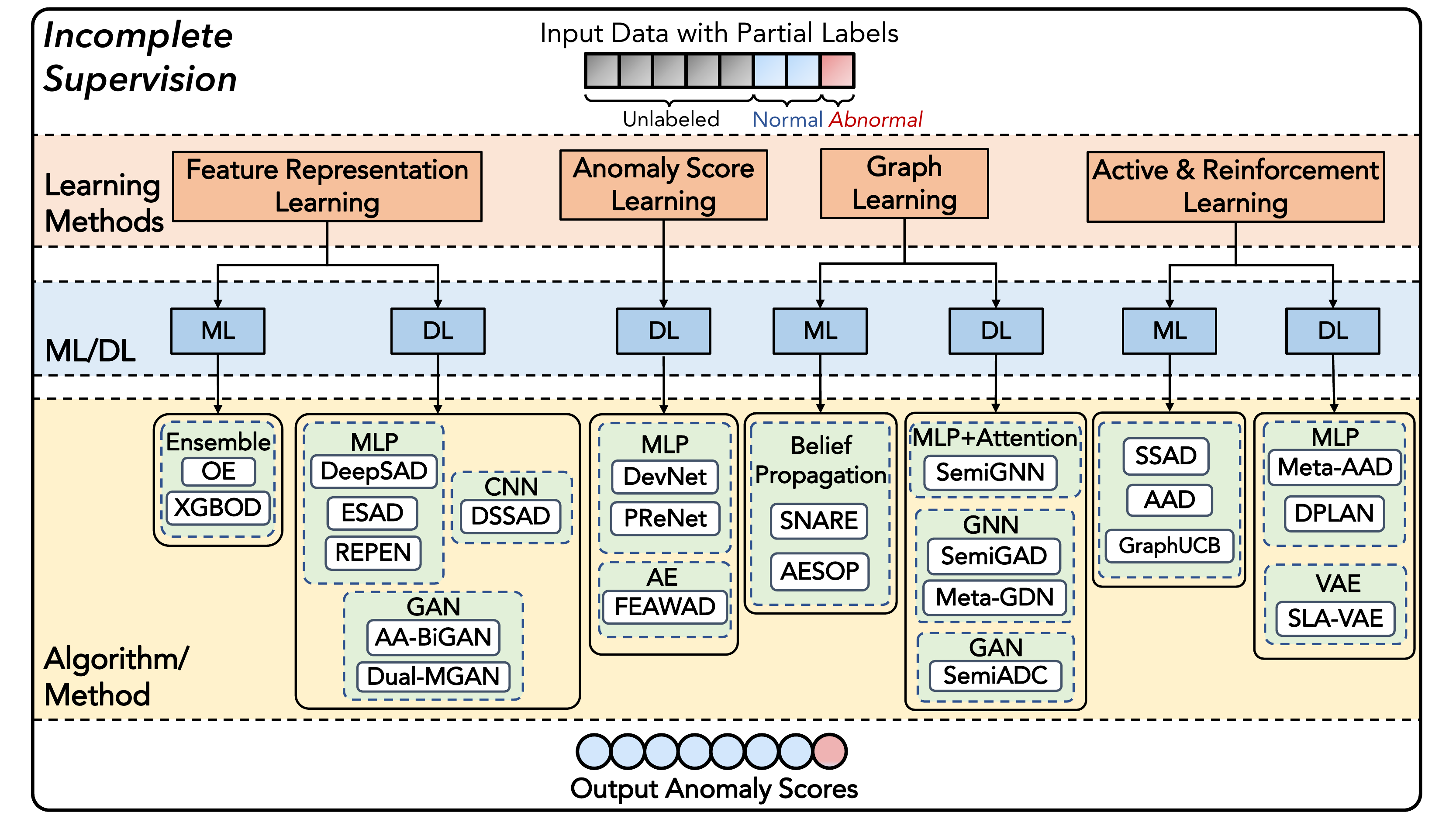}
    \caption{Category of incomplete supervision.}
    \label{fig:incomplete}
    \vspace{-0.25in}
\end{figure}

\textbf{Overview}. Fig. \ref{fig:incomplete} provides a high-level overview of the included algorithms, and we briefly describe them below with pros and cons. First, the most notable group of methods is \textit{anomaly feature representation learning}, which learns meaningful embeddings that can distinguish normal and abnormal data. 
The key strength of this group of methods is simplicity, which is primarily adapted from unsupervised representation learning (e.g., one-class or reconstruction-based method) by making minor changes to the unsupervised objective function. However, the learned normal data representation may be sub-optimal due to either the impact of unlabeled anomaly noise or the excessive learning of labeled anomalies.
Differently, \textit{anomaly score learning} directly maps the input data to anomaly scores within an end-to-end learning paradigm. On the other hand, the model training is fully guided by the known labeled data, and the model may fail to detect unknown (i.e., newly occurred) anomalies. 
Under the context of graph data, primary methods focus on \textit{label propagation}, which leverages a message-passing mechanism to propagate partial label information according to some inherent connections in the data. The additional graph structure brings additional information for detection, while posing challenges in handling the graph structure efficiently.  We conclude this section by describing \textit{active learning and reinforcement learning}, which handles incomplete supervision via getting more labels efficiently. This approach may significantly improve the performance by directly addressing the insufficiency of labels, while extra human annotation costs cannot be ignored. 
\noindent \textbf{Anomaly feature representation learning}.
% \mq{with partially labeled data?}
% \mq{anomalies?} 可能用data就行
% . Semi-supervised learning 
% is a standard approach to address incomplete supervision. 
% In a nutshell
The key idea is to improve unsupervised representation learning by partially available supervision information\footnote{We exclude the algorithms that only use normal data. Such methods have been shown inferior to those anomaly-informed AD algorithms \cite{han2022adbench}, as many of their identified anomalies are simply data noises or unimportant data objects, owing to the lack of prior knowledge of interested anomalies \cite{pang2019deepdevnet}.}. The earlier solution \cite{micenkova2014learning,zhao2018xgbod} uses unsupervised AD algorithms as feature extractors to get useful representations from the underlying data, where the anomaly scores predicted by these unsupervised AD algorithms are appended to the original input features for training a supervised classifier like XGBoost \cite{chen2016xgboost}. 
The assistance of unsupervised representation learning can be regarded as compensation for the incompleteness of labels.

% BORE \cite{micenkova2014learning} and  Extreme Gradient Boosting Outlier Detection (XGBOD) \cite{zhao2018xgbod} uses multiple  the predictive capabilities of an embedded supervised classifier on an improved feature space.
More recent works have employed \textit{deep neural networks} to extract anomaly-oriented representation from the input data, other than manually building unsupervised extractors. 
Among them, DeepSAD \cite{ruff2019deep} 
% \mq{is the most notable work?} 
improves the unsupervised deep one-class method DeepSVDD \cite{ruff2018deep} with the aid of a limited number of labeled instances. The method imposes a penalty on the inverse of the embedding distance of the anomaly feature representation, requiring anomalies to be mapped farther away from the center of an embedding hypersphere. Moreover, the paper provides insights into the connection between deep one-class learning and DeepSAD from the view of information theory.
% is the most notable work, 
% introduces semi-supervised learning to their 
% which is built on top of unsupervised work DeepSVDD \cite{ruff2018deep}. In addition to empirical analysis, the authors conduct insightful analysis on semi-supervised AD through the view of information theory. 
% They also demonstrate its power through visualizations of decision boundaries of various learning paradigms, which shows that semi-supervised AD can strike a balance between one-class learning and classification. 
Following DeepSAD, \cite{huang2020esad} proposes an encoder-decoder-encoder structure to leverage the partial label information by optimizing the mutual information and the entropy in the first and the second encoder, respectively, and enforcing consistent latent representations between the two encoders.
% \mq{to revise}.
% 下面这篇也属于对于DeepSAD的一种改进,但没那么重要,看看要不要保留
Similarly, \cite{feng2021learning} proposes a harmonious loss to coordinate the loss and gradient between unlabeled and labeled samples, which makes the training process suffer less perturbation caused by the pollution of unlabeled anomalies and magnifies the effect of labeled samples to better guide model training. Also, a transfer learning scheme is designed to enhance the expressiveness of the representations. 

In order to learn expressive feature representations of ultrahigh-dimensional input data, REPEN \cite{pang2018learning} combines representation learning and detection by using a ranking model-based approach to learn a low-dimensional embedding that is optimized for random distance-based detection methods. The labeled anomalies are served as an enhancement of the triplets for calculating the representation distance in the hinge loss function, and help the model learn more application-relevant representations.

% Different from DeepSAD and its related papers, REPEN \cite{pang2018learning} proposes a ranking model-based framework, which unifies representation learning and anomaly detection to learn low-dimensional representations tailored for random distance-based detectors. The labeled anomalies are served as an enhancement of the triplets for calculating the representation distance in the hinge loss function. \textcolor{red}{why it is different from the above group.}
% % 

% \textbf{GAN-based representation learning}. \yz{what is the key idea of using GAN?} \comment{using latent spaces in GANs as representations, said by DevNet. Not sure.} 
Another group of representation learning works uses generative adversarial networks (GAN) to enhance the model training and/or data augmentation.
% \textcolor{red}{due to which reason? what is the benefit of doing so?}
Based on the BiGAN \cite{donahue2016adversarial} structure, Anomaly-Aware Bidirectional GAN (AA-BiGAN) \cite{tian2022anomaly} leverages incomplete anomalous knowledge represented by the labeled anomalies to simultaneously learn a probability distribution for both normal data and anomalies. Notably, the distribution guarantees to assign low-density values for the collected anomalies. 
Dual Multiple Generative Adversarial Networks (Dual-MGAN) \cite{li2022dual} integrates multiple GANs to realize reference distribution construction and data augmentation for detecting both discrete and grouped anomalies.
% \\

%
% \noindent \textbf{Semi-supervised anomaly score learning.} 
\noindent \textbf{Anomaly score learning}.
% \za{I'm still unsure about these 2 sections. It may not be appropriate to merge them into \textbf{semi-supervised learning}. The content is a little bit large.}
Although learning feature representations improves downstream detection, it is an indirect learning for anomaly scores. Thus, some literature designs end-to-end anomaly score learning frameworks so that they can acquire anomaly scores directly via the trained model.
% which may results in sub-optimal performance. 

DevNet \cite{pang2019deepdevnet} is one of the leading works. It first assumes a Gaussian prior on anomaly scores of normal data to generate reference score samples. Then, the proposed deviation loss pushes the normal instances mapping around reference scores while enforcing a statistically significant deviation between the reference scores and the anomaly scores of labeled anomalies.
Following DevNet, \cite{pang2019deepprenet} defines AD as a task of predicting the relationship between pairs of instances and presents PReNet, which learns a two-stream ordinal regression neural network to output anomaly scores of randomly selected instance pairs.
Besides, Feature Encoding with AutoEncoders for Weakly supervised Anomaly Detection (FEAWAD) \cite{zhou2021feature} is the latest advancement. It improves upon DevNet by utilizing the deep autoencoder in DAGMM \cite{zong2018deep} to encode the input data and using the encoded hidden representation, reconstruction residual vector, and reconstruction errors as new representations to facilitate score learning.
% \za{This work will be added later.} 

% (Partial) 
\noindent \textbf{Graph learning via label propagation}. %In addition to the above methods that are primarily designed for tabular data, there is a special group of methods tailored for graph AD. Notably, graphs propagate their node information through their connectivity. 
% \za{Maybe \textbf{label propagation and graph learning} sound better? } 
For graph AD, another line of thought considers inherent connections in the data to propagate partial label information accordingly \cite{gao2023alleviating,gao2023addressing}.
% This special group of algorithms is consequently often tailored for graph AD.
Early works \cite{mcglohon2009snare,tamersoy2014guilt} use supervision signals as initial anomaly scores for nodes and conduct belief propagation to re-weight the anomaly scores in a probabilistic way. More recently, the message-passing scheme with graph neural networks (GNN) has become a mainstream framework to aggregate and propagate information across nodes.  Supervision signals from partial labels will guide this process through a designed loss function.
% With the advance of graph learning, recent works focus on utilizing specialized neural networks to boost information propagation in a semi-supervised way.

Specifically, \cite{wang2019semi} utilizes hierarchical attention to aggregate node-level and view-level information in a multi-view network. Their objective function combines supervised classification loss with a random-walk-based unsupervised graph loss. \cite{kumagai2021semi} uses graph convolutional networks (GCN) to propagate label information during the process of learning node embedding. Their objective function enforces GCN to learn a one-class description with a differentiable AUC loss acting as supervision. \cite{ding2021few} uses GNN as the graph encoder and deviation loss \cite{pang2019deepdevnet} as the objective function. The paper additionally proposes to apply meta-learning on some processed similar networks to incorporate more supervision.
% \za{I'm not sure where to put this work, it uses Meta-learning as an extension of semi-learning, maybe we should put it in semi-supervised learning section}
% To further realize the potential of semi-supervised learning, they additionally propose to apply meta-learning to incorporate more supervision by exploiting labeled anomalies that are accessible on some processed similar networks. Hence MAML \cite{finn2017model} is applied to initialize the model parameters through the information of gradients returned by sampled auxiliary networks. It's worth mentioning that their setting can also be viewed as an inaccurate supervision case since it utilizes labels from multiple auxiliary networks, which can be seen as noisy labels for the target network. In other words, this kind of meta-learning approach is also a possible choice for graph AD with multi-sourced inaccurate supervision.  
 \cite{meng2021semi} considers a scenario in dynamic graph AD where both incomplete and inaccurate supervision exist. The authors propose SemiADC, which learns the feature distribution of normal nodes in a GAN-based fashion, with regularization from a limited number of labeled abnormal nodes. The learned feature distribution and structure-based temporal correlations are then utilized to improve this learning process by self-training which iteratively cleans the noisy labeled nodes.
% \mq{to be categorized...GAN-based? Graph? Incomplete? Inaccurate?} \cite{meng2021semi} proposes a Semi-supervised Anomaly Detection framework for dynamic Communication networks (SemiADC). SemiADC leverages a GAN-based model to learn normal feature distribution with regularization from a small portion of labeled abnormal nodes, and improves this learning process of feature distribution by the self-training method with iteratively cleaning the noisy labeled nodes.
% \\
%

 \noindent \textbf{Active learning and reinforcement learning}.
% \za{\cite{pelleg2004active} \cite{das2016incorporating} can also be categorized in this section.}
In addition to using existing labels, some works focus on gathering more supervision.
One of the earliest works on \prob \cite{gornitz2013toward}, shows that the performance of unsupervised AD often fails to satisfy the requirement in applications, indicating the necessity of learning from partially labeled data.
% to guide the model generation. 
Meanwhile, the paper argues that \prob should ground on unsupervised learning instead of the supervised paradigm since the latter can hardly detect new and unknown anomalies. As a remedy, the authors propose an active learning strategy to automatically choose candidates for labeling. \cite{das2016incorporating} also leverages active learning. The proposed algorithm operates in an interactive loop for data exploration, which maximizes the total number of true anomalies presented to the expert under a query budget.
In \cite{huang2022semi}, the authors propose SLA-VAE, a variational autoencoder (VAE) method for multivariate time-series AD. The key difference here is it first uses semi-supervised VAE to identify anomalies and then uses active learning to update the existing model with a small set of uncertain samples.

% As a related direction to 
Relating to active learning, reinforcement learning (RL) has also been explored in \prob. RL often models \prob as a Markov Decision Process for sequential discovery of anomalies, where the ground-truth labels are provided as feedback by the expert. RL usually takes an anomaly detector as an agent, features of the current instance as state, labeling decision as action, sampling strategy as state transition function, and anomalies as a reward. Specifically, \cite{zha2020meta} proposes to extract transferable meta-features to ensure consistency in feature dimension across different datasets. This allows the trained meta-policy to be applied directly to new unlabeled datasets without the need for additional tuning. \cite{pang2021toward} proposes to simultaneously explore limited anomaly examples and rare unlabeled anomalies to extend the learned abnormality, resulting in joint optimization of both objectives. \cite{ding2019interactive} applies contextual multi-armed bandit algorithm for graph AD where each node is considered as an arm to pull. The proposed model first generates several node clusters to select a potential anomaly, and then queries the human expert for an update.
% selection strategy update.
% \\
%

% 

% \noindent \textbf{Might be incomplete \prob...}

% \cite{ioannidis2019graphsac}

\subsection{Open Questions and Future Directions}
\label{subsec.incomplete.future}

\textbf{Efficient label acquirement}. One immediate relief is to acquire more labels efficiently via active learning and human-in-the-loop. However, it remains underexplored in choosing valuable samples (i.e., anomalies) for annotations. In addition to the above methods discussed in this survey, one promising approach is to use learning-to-rank that dynamically adjusts the sample order for annotators so that anomalies are assigned a higher probability to be labeled \cite{lamba2019learning}. 

\noindent \textbf{Few-shot/meta learning}. Beyond WSL, there exist other methods that deal with incomplete supervision with limited usage in AD. Specifically, the setting of few-shot learning is more practical in AD and it deserves more investigation. Meta-learning can be applied to transfer supervision signals from similar datasets with more supervision \cite{ding2021few}, which makes it possible to incorporate more information instead of focusing on a single dataset. 
% Yet we have not seen much of their debut in anomaly detection. 
% \yz{Adding some descriptions of how to make it potentially.} 

\noindent \textbf{Balancing the weights of non-labeled and labeled data}. As shown in Eq.\ref{equ:balancing_factor}, most of the above representation learning methods' loss needs setting the balance factor \textcolor{blue}{$\alpha$} between the non-labeled and labeled data, which is an important hyperparameter here. The current approach primarily sets aside a hold-out data for setting it, while this may not be ideal for already insufficient labeled anomalies. One promising solution is to use automated ML to set it, which is discussed below.

\begin{equation}
   \mathcal{L} = \underbracket{\mathcal{L}_u(\mathbf{X})}_\textrm{non-labeled}+ \textcolor{blue}{\alpha}
   \underbracket{\mathcal{L}_l(\mathbf{X}, \mathbf{y}_p)}_\textrm{labeled}
    \label{equ:balancing_factor}
\end{equation}

% \yz{how to decide the balancing factor in loss? One solution is to learn it. AutoML?}

% \mq{Automatic model selection under limited label information scenario? (may raise conflict of interests)}

\noindent \textbf{Hyperparamter (HP) tuning and model selection}. In addition to the above balancing factor, there are more HPs to set, e.g., learning rate. Also, which learning model to use is also an important problem for AD applications with incomplete supervision. However, due to the limited number of labeled anomalies, existing supervised approaches may not be optimal. We want to bring attention to the works of unsupervised model selection \cite{zhao2021automatic} and hyperparameter tuning \cite{zhao2022towards} for AD.
\vspace{-0.1in}
\section{AD with Inexact Supervision}
\label{sec.inexact}

% \subsection{Definitions and Real-world Applications}
\textbf{AD with inexact supervision} refers to the scenario that although labels are available, they are not detailed enough to yield satisfactory detection performance. One common setting is that the provided labels are not available at the same level as outputs \cite{dahmen2021indirectly}. For instance, one does not have access to instance-level labels but only set-level labels (which indicate at least one instance in the set is anomalous) \cite{iwata2020anomaly}.
% instead of individual instances, labels can be only acquired for a set of instances, indicating that at least one instance in the set is anomalous \cite{iwata2020anomaly}.
% In tabular data, we are given labels at the set level, where each set contains a few samples. 
Let us see a more concrete example in financial fraud detection, where companies can obtain user-level abnormal labels from regulatory agencies or third parties. These user-level labels only indicate whether a user has engaged in illegal activities such as fraud, money laundering, or gambling, while we are more interested in which transactions are anomalous (i.e., transaction-level results). 
% For time-series data, we are often given anomaly labels per time window or series, while the abnormality of data points at each timestamp is needed as outputs.
In time-series data, anomaly labels may be given per time window or series, but the goal is to identify point-level abnormalities. 
This setting is also common for video AD, where only coarse-grained video-level labels are given but our goal is to identify segment or frame-level anomalies \cite{sultani2018real}. 
% \mq{here we mentioned that AD with inexact supervision could occur in tabular, time-series and video, but we mainly talk about the last in the following paper.}
% we may consider it closely related to the time series setting where a video is a set of images by the timestamp. In this scenario, 
% there are coarse-grained video-level labels in training data and the aim is to train a model to recognize segment or frame-level anomalies.

% Many applications can be viewed as AD with inexact supervision. 
% % \textbf{Video Anomaly Detection.} 
% For instance, 
% % Video anomaly detection is one of the most typical scenarios for AD with inexact supervision. 
% The training data for video AD may only contain video-level anomaly labels, while the goal is  to identify whether an anomaly exists at the frame level or video segment level. 
% \textbf{Fraud Anomaly Detection.}  

\begin{problem}
[AD with Inexact Supervision]
\textit{\em Given} an anomaly detection task with input data $\mathbf{X} \in \mathbb{R}^{n\times d}$
(e.g., $n$ samples and $d$ features) that can be further grouped into $k$ groups, i.e., $\{\mathbf{X}_G^1, \ldots, \mathbf{X}_G^k$\}. With the coarse-grained labels at the group level (i.e., $\mathbf{y}_{G} \in \mathbb{R}^{k}$ where $k \leq n$), 
train a detection model $M$ with $\{\mathbf{X}, \mathbf{y}_{G}\}$. Output the sample-level anomaly scores on the test data $\mathbf{X}_\text{test} \in \mathbb{R}^{m\times d}$, i.e., $\mathbf{O}_\text{test} := M(\mathbf{X}_\text{test}) \in \mathbb{R}^{m\times 1}$.

% \textcolor{blue}{define a coarse-grained label that?}
% labels $\mathbf{y}_{a}\mathbb{R}^{p}$ where $p < n$, 
% build a detection model $M$ to leverage $\{\mathbf{X}, \mathbf{y}_{p}\}$ to achieve the best performance.
\label{problem:inexact}
\end{problem}

\subsection{Key Algorithms}

\begin{figure}[!t]
    \centering
    \includegraphics[width=8.6cm]{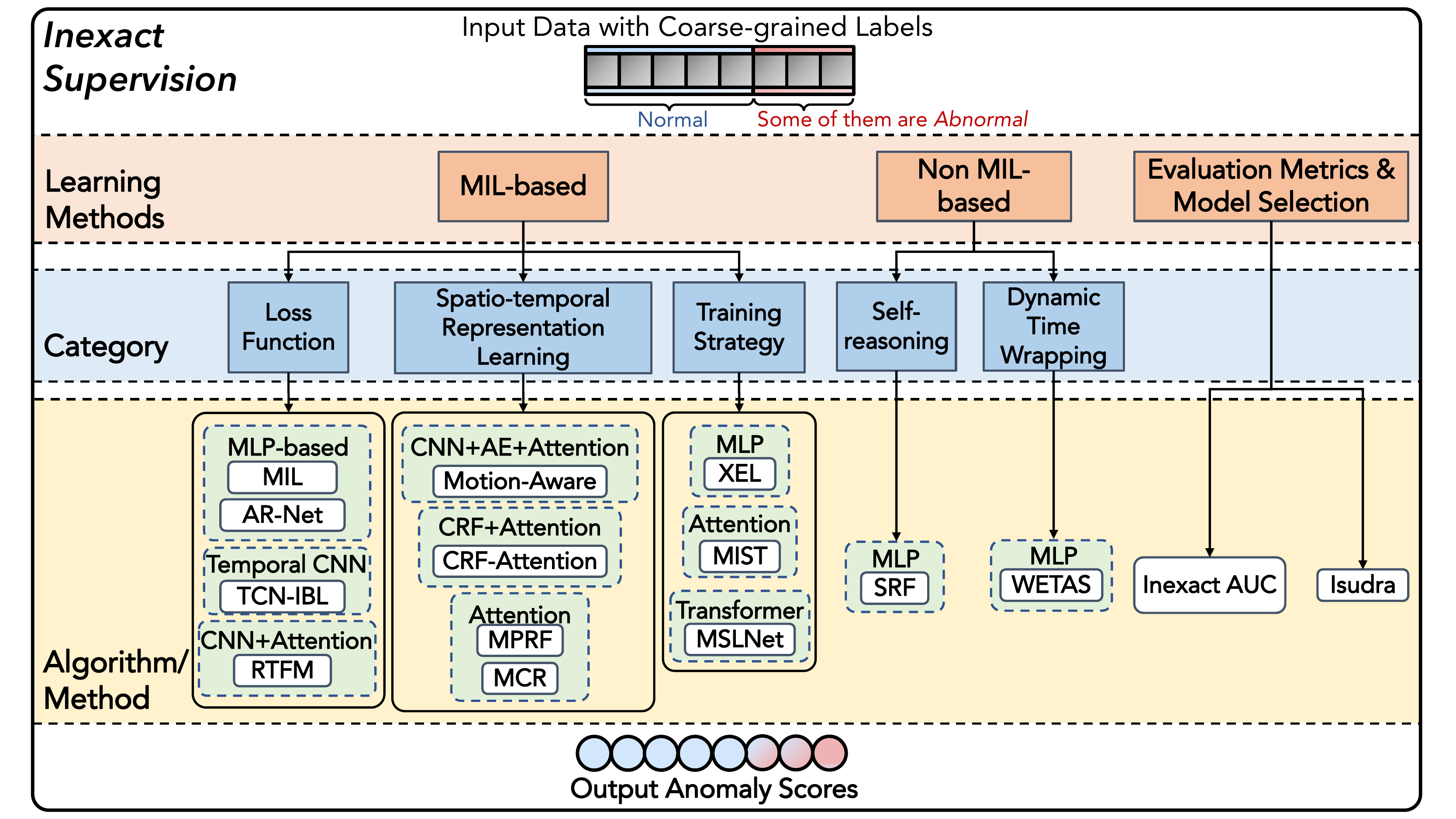}
    \vspace{-0.25in}
    \caption{Category of inexact supervision.}
    \label{fig:inexact}
    \vspace{-0.2in}
\end{figure}

\textbf{Overview}. Although AD with inexact supervision has many important applications across modalities, existing literature mainly focuses on video data. Fig. \ref{fig:inexact} provides a high-level overview of the included algorithms, where the main difference between these algorithms is whether they are based on multi-instance learning  (MIL). Beyond the scope of algorithm designing, some works propose new evaluation metrics or model selection strategies to address the problem. 
% \mq{to revise.}
% \textcolor{red}{to revise. add a description for subsection New evaluation metrics and model selection}

% \textbf{Image and Video}
% \label{subsec:inexact.image_video}

%Based on this definition, the previous work \cite{cheng2015video,luo2017revisit,liu2018future} based on the unitary classification paradigm \cite{wan2020weakly,zhong2019graph} only use training samples with normal coarsed-grained level labels to model the common pattern \textcolor{red}{shall we cite so many papers about unitary classification?}\textcolor{blue}{already removed}, and detected abnormal behavior patterns that were different from normal. However, these methods cannot effectively learn all the normal behavior patterns, and the model without guidance on abnormal patterns is difficult to achieve good results. \yz{How does the coarsed-grained level is used?}\textcolor{blue}{done}

% 画图介绍MIL？画法发展历程(信息量不大)？
% 或者树状图(参考“Time Series Data Augmentation for Deep Learning: A Survey” Figure 1)
%\textbf{Multi-Instance Learning and its variants}.
\noindent \textbf{Multi-instance learning methods}. Multiple-instance learning (MIL) is originally designed for predicting the label of a bag/set of instances, rather than the label of individual instances \cite{carbonneau2018multiple}. \cite{sultani2018real} first considers solving this AD problem via the MIL framework, where normal and abnormal videos are regarded as negative and positive bags, respectively, and video segments as instances. 
%By enforcing ranking only on the two instances having the highest anomaly score respectively in the positive and negative bags, they realize anomaly score ranking in the segment level. 
The two instances with the highest anomaly score in the positive and negative bags are considered to represent these two bags. By enforcing ranking only on the two instances, it realizes anomaly score ranking at the segment level. 
Note that only video-level coarse-grained labels are used in the model training. Moreover, it introduces both sparsity and temporal smoothness constraints in the objective function to better localize anomalies during optimization. 
Based on this work, subsequent studies focus on improving MIL in terms of several aspects, including loss function, feature representation learning, and training strategies, so as to better detect anomalies with only coarse-grained video labels.
% Based on this work, many works is proposed to detect anomalies better with coarse-grained labels, and most of them are mainly improved in terms of loss function and network architecture.

There exist some works that \textit{modify the objective function} in the original MIL framework. 
% loss
\cite{zhang2019temporal} employs a temporal convolutional network (TCN) to model the temporal structure in the video and proposes to further constrain the large %searching 
search space in the MIL framework, which assumes that the gap between the lowest score (i.e., most likely to be the normal instance) and the highest score (i.e., most likely to be the abnormal instance) of input instances in a positive bag should be large, whereas that in the negative bag should be small.
% 
%\cite{wan2020weakly} proposes the Anomaly Regression Net (AR-Net), which includes both the dynamic multi-instance learning loss and the center loss, where the former is used to enlarge the inter-class distance between anomalous and normal instances, while the latter is proposed to reduce the intra-class distance of normal instances.
% Rephrase:
Anomaly Regression Net (AR-Net) \cite{wan2020weakly} views the problem as an anomaly score regression task, which includes two specific loss functions. The dynamic multi-instance learning loss here is designed to increase the difference between anomalous and normal instances while the center loss can decrease it within normal instances.

Robust Temporal Feature Magnitude (RTFM) \cite{tian2021weakly} enhances the robustness of MIL by considering the top-$k$ instance (instead of top-$1$ in the original MIL) within a video, and replacing the instance anomaly scores with learning feature magnitude to more effectively recognize the abnormal video segments. 
%RTFM further adapts dilated convolutions and self-attention mechanisms to capture long- and short-range temporal dependencies to enhance the learning process of the feature magnitude. 
Further, the dilated convolutions and self-attention mechanisms are utilized in RTFM to enhance the learning process of the feature magnitude by capturing both long- and short-term temporal dependencies.

% feature representation learning
Considering spatio-temporal characteristics (i.e., the object motion and temporal consistency) in the video segments, several works also discuss how to extract more meaningful features for improving the downstream AD with inexact labels.
\cite{zhu2019motion} improves the MIL framework by obtaining motion-aware features and capturing the temporal context in the video segments. The proposed temporal MIL ranking loss is calculated between the normal and anomalous videos' overall anomaly scores, which are the weighted sum of the anomaly scores of individual instances by the learned attention weights. 
% Besides, the temporal smoothness constraint introduced in the original MIL framework is removed as they empirically find it harmful for model training.
% 
% \mq{to do mark} In the previous work of weakly supervised video anomaly detection, pre-training models such as I3D and C3D are usually used to capture video features. 
In \cite{purwanto2021dance}, a relation-aware feature extractor is proposed to capture the multi-scale CNN features, and a self-attention that is integrated with conditional random fields (CRFs) is developed to learn both short-range correlations and inter-dependency of these multi-scale CNN features. This work still follows the MIL framework, while adding a comparative loss to further expand the gap between positive and negative instances. 
\cite{wu2021weakly} introduces a weakly supervised spatio-temporal AD in surveillance video, which aims to detect the fine-grained spatio-temporal location of abnormal behaviors, predicting not only when but also where the anomaly happens in the video. This work proposes a dual branch network to extract two kinds of tube-level instance proposals and employs a relationship reasoning module to capture the correlation between tubes/videolets. Further, 
%they 
it introduces a framework called ``Mutually-guided Progressive Refinement'' to progressively refine the corresponding branch in a recurrent manner.
%Mutually-guided Progressive Refinement framework is set up to employ dual-path mutual guidance in a recurrent manner, progressively refining the corresponding branch and the whole framework \textcolor{red}{need to check this sentence. What is this for...}.

More recently, \cite{gong2022multi} proposes a Multi-scale Continuity-aware Refinement network (MCR) to improve the MIL framework. 
%MCR extracts the continuity of instances in different temporal scales by using different window sizes to compute the anomaly score vectors, which are further combined by the learned attention weights.
%phrase:
MCR uses different window sizes to calculate the anomaly score vectors for instances at various temporal scales, thus capturing the continuity over time. These vectors are then combined using attention weights that have been learned during the training process.

Other studies focus on applying different \textit{model training strategies} to enhance MIL.
\cite{yu2021cross} collects the hard normal instances in the previous training epochs to fulfill a cross-epoch learning, where the validation loss is proposed to suppress the anomaly scores of these hard normal instances. Furthermore, 
%they propose
it proposes a dynamic margin loss function to realize the anomaly score margin between the hard normal instances and the most deviant instances in abnormal bags, where the margin is progressively increasing during the training process. 
\cite{feng2021mist} introduces a multiple-instance self-training framework (MIST), which generates clip-level pseudo labels with (1) a strategy for selecting sparse, continuous samples to enable the generator to understand the context surrounding anomalies and (2) a self-guided feature encoder that utilizes attention mechanism to focus on anomalous regions and generate task-specific representations.
% Notably, the former guides the generator to learn the context around anomalies, and the latter emphasizes the anomalous regions as extracting task-specific representations.
% is developed, which is composed of a clip-level pseudo label generator with a sparse continuous sampling strategy that guides the generator to learn context around anomalies,
% and a self-guided attention boosted feature encoder to 
% 
\cite{li2022self} proposes a Multi-Sequence Learning (MSL) to reduce the error probability of the selected top abnormal snippet in the typical MIL framework. The learned snippet-level anomaly scores in MSL can be gradually refined by a self-training strategy, where their fluctuations are effectively suppressed by the predicted video-level anomaly probability in the inference stage.

\noindent \textbf{Learning methods beyond MIL}.
% Unlike those methods, some works focuses more on the way of model training to enhance the AD performance under inaccurate supervision. 
Although the majority of the works focus on MIL, there are other approaches to address the problem. For instance, \cite{zaheer2020self} proposes to train neural networks in a self-reasoning manner with inexact video-level labeled data. This is achieved by generating pseudo labels through binary clustering of 
%spatio-temporal video features extracted by a pre-trained model
video features with both spatial and temporal information extracted by a pre-trained model, and ensuring consistency between these generated labels and those predicted by the network.
WEak supervision for Temporal Anomaly Segmentation (WETAS) \cite{lee2021weakly} leverages the video-level (i.e., weak) anomaly labels to infer the sequential order of normal and anomalous segments as a rough segmentation mask, where the dynamic time wrapping (DTW) alignment between the input instances and this estimated segmentation mask is used for obtaining the temporal segmentation.

%\cite{georgescu2021anomaly}
%没有任何涉及弱监督，或者利用粗粒度标签信息的介绍

\noindent \textbf{Labels for evaluating and selecting unsupervised methods}. Other than designing new detection methods, \cite{iwata2020anomaly} introduces a new evaluation metric for the performance measurement with coarse-grained labels called \textit{inexact AUC (i.e., the area under the ROC curve)}. Based on \textit{inexact AUC}, it 
proposes a reconstruction-based AD method that minimizes the reconstruction error while maximizing \textit{inexact AUC}. Further, it mentions that the proposed approach suits any unsupervised model in addition to the autoencoder utilized in this paper, which brings an interesting angle to be further investigated.
Differently, \cite{dahmen2021indirectly} uses the labels for model selection on unsupervised AD methods, where the approach is named ``indirectly supervised AD''. 
%design a reconstruction-based AD method to maximize it. 
% \vspace{-0.15in}

% % \textbf{Time-series}
% % \label{subsec:inexact.ts}

% \noindent \textbf{Model selection}. Other than directly using coarse-grained labels to train a detection model, 

% \cite{iwata2020anomaly}

% \cite{hou2021stock}

% \subsection{Graph}
% \label{subsec:inexact.graph}

\subsection{Open Questions and Future Directions}
\label{subsec.inexact.future}

\textbf{Call for research on other modalities with inexact supervision}. 
Notably, the above literature is mostly for videos, while inexact supervision has not been widely studied for the tabular, time-series, and graph data. For example, financial companies may have access to user-level fraud labels from regulatory agencies, while they need to detect transaction-level fraud instances, which could be either time-independent (i.e., tabular data) or time-continuous (i.e., time-series).
% Inexact scenarios in graph AD is a yet-to-explore direction with numerous applications. 
For graphs, the subgraph labels are available while the goal is to identify node-level anomalies in a graph \cite{liu2022bond}. Another interesting setting is that we have node-level labels, but need to perform edge-level detection. For example, we know a fraudster (as a node) in a transaction network, but can not locate his/her specific fraudulent transactions (as edges). 
% Some edge detection methods may be helpful under this setting.
% For example, with a transaction network, we want to figure out anomaly nodes but we only have labels for a few node clusters --- these labels are not granular enough at our task level. \yz{what is a potentially good solution?}. 
For these underexplored problems, one may consider adapting the MIL techniques from the above literature.

% \za{Maybe a more interesting scenario exclusive for graph AD  }
% \textcolor{red}{Can you turn this into a short paragraph? We need at least two directions per setting.}
% \mq{apply the existing method of inexact supervision to time-series domain, or use time-series technique to extract temporal information of video.}
\noindent \textbf{Cross-modality methodology adaptation}. 
% \za{This "Technology transfer" sounds weird. May be "\textbf{Communication between inexact time series and vedio methods}" ?} 
 % This leaves opportunities to transfer the video AD techniques for inexact supervision to these tasks. 
 Also, different data modalities share characteristics, e.g., both time series and videos have time stamps and temporal relationships.
 Thus, we may transfer leading techniques from time-series to video AD and vice versa, e.g., AD-specified transformer architecture or attention mechanism \cite{tuli2022tranad,zhou2022fedformer}, could also be served as an effective solution to better capture the temporal information in the video.
 \vspace{-0.2in}
\vspace{-0.1in}
\section{AD with Inaccurate Supervision}
\label{sec.inaccurate}

% \subsection{Definitions and Real-world Applications}
\vspace{-0.05in}
\textbf{AD with inaccurate supervision} considers the situation where the labels are noisy and corrupted, which is also common in AD applications where accurate annotation is expensive.
% , especially in security and finance.
Using network intrusion detection as an example, obtaining accurate labels can be difficult due to the sensitive nature of the data and the expense of annotation. However, there is often a significant amount of weak or noisy historical security rules available for detecting intrusion from various angles such as unauthorized network access and unusual file movement \cite{zhao2022admoe}.
Though not as perfect as human annotations, these noisy sources  
generated by human annotators or machine detectors/rules  
are valuable as they encode prior detection knowledge, and thus are used as ``inaccurate/weak'' labels. 
% This reasoning applies to many important fields in security and finance. 
% Clearly, the most straightforward way to the problem is to fix the corruption in the label, which is widely known as 
Clearly, this problem is a sub-problem of noisy label learning \cite{frenay2013classification}, while AD faces additional challenges like data imbalance that existing methods cannot handle \cite{zhao2022admoe}.

\begin{problem}
[AD with Inaccurate Supervision]
\textit{\em Given} an anomaly detection task with input data $\mathbf{X} \in \mathbb{R}^{n\times d}$
(e.g., $n$ samples and $d$ features) and noisy/inaccurate labels $\mathbf{y}_{w} \in \mathbb{R}^{n}$, 
% build a detection model $M$ to leverage $\{\mathbf{X}, \mathbf{y}_{w}\}$ to achieve the best performance.
train a detection model $M$ with $\{\mathbf{X}, \mathbf{y}_{w}\}$.
Output the anomaly scores on the test data $\mathbf{X}_\text{test} \in \mathbb{R}^{m\times d}$, i.e., $\mathbf{O}_\text{test} := M(\mathbf{X}_\text{test}) \in \mathbb{R}^{m\times 1}$.
\label{problem:inaccurate}
\end{problem}
% \vspace{-0.1in}

% \subsection{Tabular}
% \label{subsec:inaccurate.tabular}
\subsection{Key Algorithms}
% \textbf{Tabular}

\begin{figure}[!t]
    \centering
    \includegraphics[width=8.6cm]{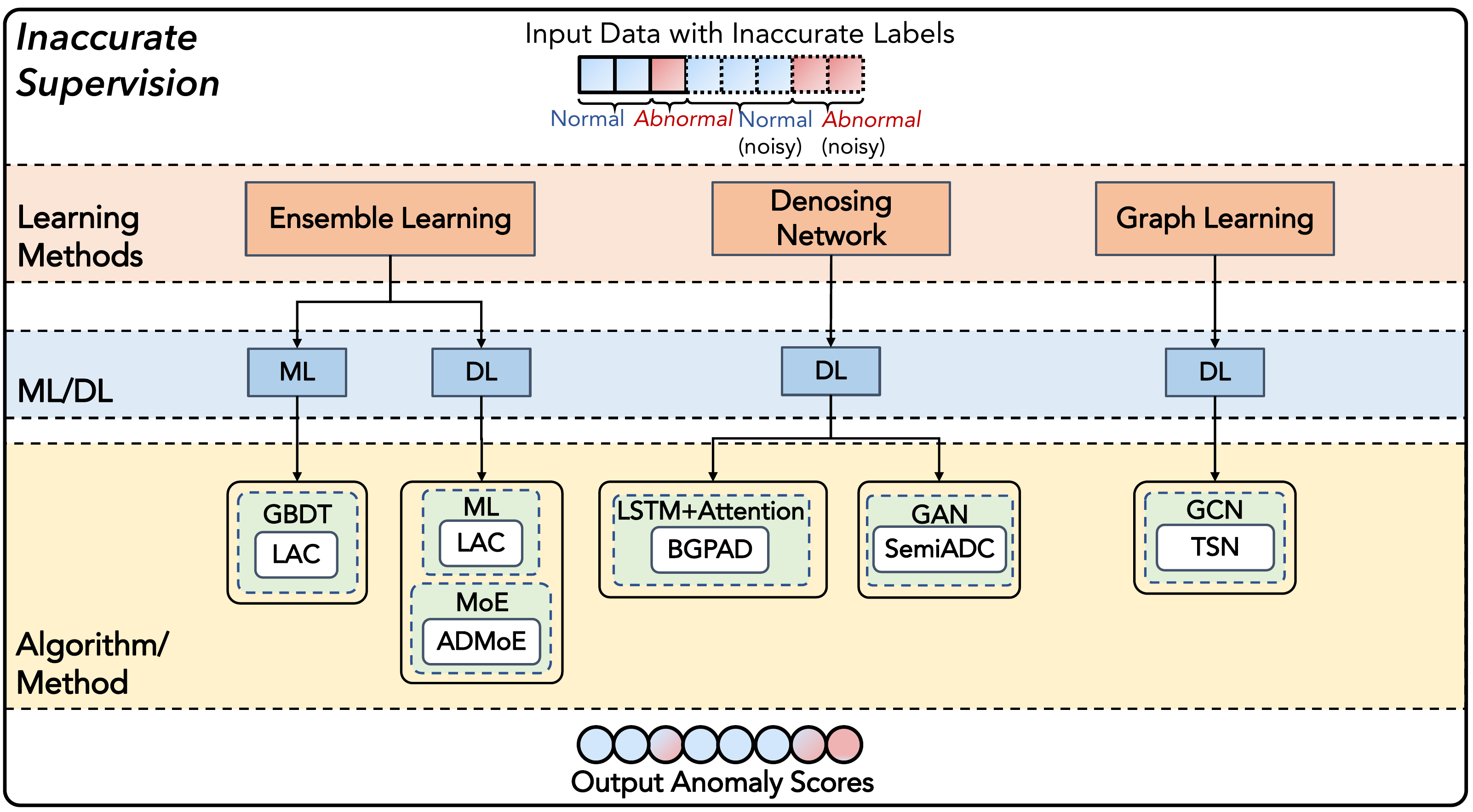}
    \vspace{-0.2in}
    \caption{Category of inaccurate supervision.}
    \label{fig:inaccurate}
    \vspace{-0.15in}
\end{figure}

\textbf{Overview}. There is only a small set of works for AD with inaccurate supervision. Fig. \ref{fig:inaccurate} provides a high-level overview of the included algorithms. Existing studies manage to extract useful label information from multiple noisy sources by leveraging ensemble learning and crowdsourcing techniques, or designing denoising networks that minimize the impact of label noises in the model training process. Besides, there exist few works that provide solutions for other \prob problems from the perspective of inaccurate supervision.

% \textcolor{red}{add a brief description for each subsection here.}

\noindent \textbf{Ensemble learning and crowdsourcing for multiple sets of inaccurate/noisy labels}. The primary approach for inaccurate supervision in AD is to learn a joint model from multiple sets of noisy labels. For instance, Label Aggregation and Correction (LAC) \cite{zhang2021fraud} proposes a two-stage method for fraud detection. 
The label aggregation stage integrates multi-sourced extremely noisy labels into aggregated labels via weighted voting, then the label correction stage corrects the aggregated labels with the aid of some  accurately labeled instances to train a robust detector.
Under a similar setting but taking an end-to-end approach, ADMoE \cite{zhao2022admoe} leverages a mixture-of-experts architecture to encourage specialized and scalable learning from multiple noisy sources. It captures the similarities among diverse sets of noisy labels via sharing most model parameters while encouraging specialization by building ``expert'' sub-networks. This parameter-sharing mechanism may be considered as ``aggregation'', while the expert specialization as ``correction''.

\noindent \textbf{Denoising networks}. Differently, selected works focus on reducing label noise. For instance, \cite{dong2021isp} proposes a WSL framework to detect Border Gateway Protocol (BGP) anomalies in the network. It uses a self-attention based LSTM 
% as a de-noising detection model 
to minimize the impact of data noise in the inaccurately labeled datasets constructed by a knowledge distillation of AD systems, where existing AD systems are used as ``teachers''. \cite{meng2021semi} proposes a GAN-based denoising network for dynamic graph AD which applies self-training to iteratively clean inaccurate node labels, as mentioned in \S \ref{sec.incomplete}. 
% \za{Added Semi ADC}

\noindent \textbf{Handling inaccurate supervision in videos}. Instead of regarding the weakly supervised video anomaly detection as an inexact supervision problem (i.e., using the coarse-grained video-level labels mentioned in \S \ref{sec.inexact}), \cite{zhong2019graph} considers it as supervised learning with noisy labels, where both feature similarity and temporal consistency graph modules are proposed to correct noisy labels. It proposes a loss function based on the direct supervision that computes a cross-entropy error over the high-confidence instances, and the indirect supervision that uses a temporal-ensembling strategy \cite{laine2016temporal} to improve the training process.

\subsection{Open Questions and Future Directions}
\label{subsec.inaccurate.future}

\noindent \textbf{Call for research on AD with inaccurate supervision}. This survey reveals that there are only limited works on this important topic. Among all, ensemble learning with multiple sets of noisy labels appears to be effective and accessible in real-world applications. 
% it is known to us that the majority of labels come from human annotators or machine detectors/rules, which are hardly infallible as the amount of data is more and more.
We thus call for more research in considering the existence of multiple sets of noisy labels--one may adapt existing multiple-set noisy-label learning methods \cite{frenay2013classification} for AD.

\noindent \textbf{Leveraging self-supervised learning in denoising labels}. Recent advancement of self-supervised learning (SSL) \cite{li2021cutpaste} may lead to a promising approach to address inaccurate labels via exploring intrinsic knowledge from the data. We thus want to mention the potential of using SSL in reducing noise in the AD labels or working as a ``regularizer".
\vspace{-0.15in}
\vspace{-0.1in}
\section{Experiments with Incomplete Supervision}

\noindent \textbf{Motivation}. Due to its popularity and importance, we present the experiments for key incomplete supervision algorithms introduced in \S \ref{sec.incomplete}, under different levels of supervision.

\noindent \textbf{Experiment settings}. In total, we benchmark 6 algorithms against 47 datasets. Algorithm implementations are based on our prior works, e.g., Python Outlier Detection (PyOD) \cite{zhao2019pyod} and Anomaly Detection Benchmark (ADBench) \cite{han2022adbench}. All the code and datasets are released.

\begin{table}[!t]
\scriptsize
\caption{Detection performance of selected incomplete supervision \prob  algorithms on 47 tabular datasets with varying labeled anomaly ratio $\gamma_{l}=1\%, 5\%,25\%$ and $50\%$.}
\vspace{-0.1in}
\label{experiment_incomplete}
\begin{subtable}{1\linewidth}
\centering
\caption{AUC-ROC results of model comparison.}
\vspace{-0.2cm}
        \begin{tabular}{lcccc}
        \toprule
        \textbf{Model} & $\gamma_{l}=1\%$ & $\gamma_{l}=5\%$ & $\gamma_{l}=25\%$ & $\gamma_{l}=50\%$ \\
        \midrule
        XGBOD & 80.03  & 86.68  & 93.20  & 95.28  \\
        DeepSAD & 75.25  & 81.74  & 89.64  & 92.72  \\
        REPEN & 77.20  & 82.23  & 86.26  & 87.45  \\
        DevNet & 79.05  & 85.94  & 89.76  & 90.97  \\
        PReNet & 79.04  & 85.66  & 89.88  & 91.11  \\
        FEAWAD & 73.93  & 82.44  & 89.20  & 91.55  \\
        \bottomrule
        \end{tabular}

\vspace{0.2cm}
\caption{AUC-PR results of model comparison.}
% \vspace{-0.2cm}
        \begin{tabular}{lcccc}
        \toprule
        \textbf{Model} & $\gamma_{l}=1\%$ & $\gamma_{l}=5\%$ & $\gamma_{l}=25\%$ & $\gamma_{l}=50\%$ \\
        
        \midrule
        XGBOD & 46.23  & 61.58  & 75.89  & 80.57  \\
        DeepSAD & 38.06  & 49.65  & 67.04  & 74.47  \\
        REPEN & 46.57  & 56.38  & 63.39  & 65.73  \\
        DevNet & 53.61  & 64.01  & 69.52  & 71.13  \\
        PReNet & 54.52  & 64.19  & 70.46  & 71.62  \\
        FEAWAD & 51.19  & 62.30  & 69.65  & 72.34  \\
        \bottomrule
        \end{tabular}
\end{subtable}%
\vspace{-0.15in}
\end{table}
\noindent \textbf{Results}. Table~\ref{experiment_incomplete} reports the model performance of AUC-ROC and AUC-PR in incomplete supervision setting.
First, AD models generally perform poorly under very limited supervision, say 1\% labeled anomalies. Even for the two best models XGBOD (80.03\% for AUC-ROC) and PReNet (54.52\% for AUC-PR), there is still great room for improvement. 
Second, we indicate that some labeling techniques like active learning can be served as an effective solution as mentioned in \S \ref{subsec.incomplete.future}, since all \prob methods yield  significant improvement when (slightly) more supervision is obtained (i.e., from $\gamma_{l}=1\%$ to $5\%$).
Also, there is no universal model that fits all situations, which agrees with our suggested automatic model selection strategies in \S \ref{subsec.incomplete.future}. See more in our repo.
% \mq{add more experiments to enrich this section, may add Dual-MGAN, and exclude AA-BiGAN since it does not release the version for tabular data}

% 不同labeled anomalies最好的model不同,model selection
% 1%的情况下如果很差，incomplete supervision的场景还是非常困难...balabala,但如果5%效果很好,或许可以利用上active learning,对于实际场景有很大提升
% 1% ~ 100%?

\noindent \textbf{Code and resources}. 
% To foster future research on this important topic, 
We open-source the above experiments and datasets at \url{https://github.com/yzhao062/wsad}, along with the collected code of surveyed \prob methods.
\vspace{-0.1in}
\section{Discussions and Concluding Remarks}
\vspace{-0.05in}
% 不同分类WSAD的转换
% In this survey, we present the first systematic analysis of weakly supervised anomaly detection (\prob) problems and methods, which concern detection tasks with incomplete, inaccurate, and inexact labels. Among the surveyed algorithms, most methods focus on incomplete supervision due to the high cost of annotation, while we argue the other two settings are equally important and call for more research.
In this survey, we present the first comprehensive analysis of weakly supervised anomaly detection (\prob) problems and methods. \prob concerns detection tasks with labels that are incomplete, inaccurate, or inexact. The majority of surveyed algorithms focus on incomplete supervision due to the high cost of annotation. However, we argue that the other two settings, inaccurate and inexact supervision, are equally important and call for further research in these areas.

% Additionally, 
We also want to highlight two interesting phenomena in \prob. Firstly, these three weak supervision problems often occur simultaneously \cite{meng2021semi}. 
% For instance, inexact labels are possibly inaccurate/noisy, and thus new techniques for joint optimization can be useful. However, addressing inaccurate labels can be increasingly hard with a small set of incomplete labels. 
For example, inexact labels can also be inaccurate or noisy---new techniques for joint optimization can be useful. However, addressing inaccurate labels can be increasingly difficult with a small set of incomplete labels.
Second, different types of weak supervision can be ``convertible". For example, inexact supervision can be regarded as a special case of inaccurate supervision at the detection level. Also, incomplete supervision may be viewed as inaccurate supervision, with the unlabeled samples being noisy. Thus, one may adapt existing algorithms from one setting to another by considering different formulations.

% \subsection{Conversion in different settings}
% \comment{some papers\cite{wan2020weakly,zhong2019graph} hold that models that only use normal training samples to learn the common pattern, and detected different behavior patterns as abnormal are also weakly supervised model, is that right?}

% Another interesting observation is that not all the existing methods focus on designing new detection algorithms, but attempt to enrich labels via active learning, denoise labels, or use limited labels for model selection, instead. Although these approaches do not directly improve detection algorithms, they bring us interesting angles to rethink \prob.

% \mq{One possible future direction is that we find there exist few papers focusing on automatic model selection in the weakly-supervised AD scenario, though that in unsupervised AD problem has been studied}

% \mq{another possible future direction is to transform the problem (e.g., inexact to inaccurate supervision), therefore we can transfer some techniques to the target task.}

% \mq{another possible future direction is to study the label propagation in the existing tabular-based weakly-supervised AD methods, i.e., the weakly-supervised LUNAR, using graph structure to propagate label information from the existing known label data to unlabeled data}

\clearpage
\newpage

% \section*{Ethical Statement}

% There are no ethical issues.

% \section*{Acknowledgments}

%% The file named.bst is a bibliography style file for BibTeX 0.99c
\bibliographystyle{named}
\bibliography{ijcai22}

\end{document}